\documentclass{article}

\usepackage{graphicx} 
\usepackage{booktabs}
\usepackage{cite}

\usepackage[T2A]{fontenc}
\usepackage[utf8]{inputenc}
\usepackage[english, russian]{babel}
\usepackage{hyperref}
\hypersetup{
    colorlinks=true,
    linkcolor=blue,
    citecolor=blue,
    urlcolor=blue,
}

\title{\textbf{AINL-Eval 2025 Shared Task: \\Detection of AI-Generated Scientific Abstracts in Russian}}
\author{\textbf{Tatiana Batura}\textsuperscript{1}, \textbf{Elena Bruches}\textsuperscript{1,2}, \and \textbf{Milana Shvenk}\textsuperscript{2}, \textbf{Valentin Malykh}\textsuperscript{3} \\
\\
\small \textsuperscript{1}A.P. Ershov Institute of Informatics Systems, Novosibirsk, Russia \\
\small \textsuperscript{2}Novosibirsk State University, Novosibirsk, Russia \\
\small \textsuperscript{3}ITMO University, Saint Petersburg, Russia
}

\date{}

\begin{document}

\maketitle

\selectlanguage{english}
\begin{abstract}
    The rapid advancement of large language models (LLMs) has revolutionized text generation, making it increasingly difficult to distinguish between human- and AI-generated content. This poses a significant challenge to academic integrity, particularly in scientific publishing and multilingual contexts where detection resources are often limited. To address this critical gap, we introduce the AINL-Eval 2025 Shared Task, specifically focused on the detection of AI-generated scientific abstracts in Russian. We present a novel, large-scale dataset comprising 52,305 samples, including human-written abstracts across 12 diverse scientific domains and AI-generated counterparts from five state-of-the-art LLMs (GPT-4-Turbo, Gemma2-27B, Llama3.3-70B, Deepseek-V3, and GigaChat-Lite). A core objective of the task is to challenge participants to develop robust solutions capable of generalizing to both (i) previously unseen scientific domains and (ii) models not included in the training data. The task was organized in two phases, attracting 10 teams and 159 submissions, with top systems demonstrating strong performance in identifying AI-generated content. We also establish a continuous shared task platform to foster ongoing research and long-term progress in this important area. The dataset and platform are publicly available at: \url{https://github.com/iis-research-team/AINL-Eval-2025}
\end{abstract}


\section{Introduction}

In recent years, the development of large language models (LLMs) has revolutionized natural language processing. These models are now capable of generating text that closely resembles human writing, making it increasingly difficult to distinguish between AI and human-generated content. This capability has led to numerous applications across various domains, with researchers proposing LLM-generated paper texts~\cite{Ayache2022Generating, chen-etal-2021-scixgen-scientific}, academic posters~\cite{Xu2021Neural}, and even research ideas~\cite{Gu2024Interesting, Radensky2024Scideator, Si2024Can}. Although this represents significant technological progress, there are domains in which the unrestricted use of LLM raises concerns, particularly in scientific publishing and academic integrity. In response to these challenges, several detection tools have emerged, such as LLM-DetectAIve~\cite{abassy-etal-2024-llm} and M4~\cite{wang-etal-2024-m4}, designed to identify AI-generated content.

The scientific community faces a growing challenge as AI-generated papers become more sophisticated and harder to detect. This is particularly concerning in multilingual contexts where detection tools and evaluation benchmarks may be less developed for languages other than English. To address this gap, we introduce AINL-Eval 2025, a shared task focused on detecting AI-generated scientific abstracts in Russian. Unlike previous efforts such as the RuATD Shared Task 2022~\cite{Shamardina2022FindingsOT}, which addressed AI-generated texts across multiple domains, including machine translation and paraphrase generation, our task focuses specifically on scientific texts in Russian, creating a specialized testbed for academic content integrity. 

In this paper, we introduce a new dataset of scientific abstracts in Russian to distinguish between human- and AI-generated texts and design challenges that require participants to develop solutions capable of generalizing to new domains and detecting texts generated by models not included in the training data. Additionally, we have created a continuous shared task platform that remains accessible for community contributions and supports long-term progress in this important area.

\section{Background}

Generally, methods for detecting AI-generated text fall into three categories: watermarking techniques, statistical approaches, and machine learning-based methods.

\textbf{Watermarking} is a technique designed to incorporate robust detection signals into machine-generated text ~\cite{pmlr-v202-kirchenbauer23a}. A watermarking method typically consists of three components: watermark, encoder, and decoder. An encoder \texttt{E} embeds a watermark into a content, while a decoder \texttt{D} decodes a watermark from a content (watermarked or unwatermarked). When a content has the watermark \texttt{w}, the decoded watermark is similar to \texttt{w}. Watermarking can be implemented using hand-crafted heuristics ~\cite{bi-2007-robust, wen-tree-2023} or using neural networks-based methods \cite{fernandez-stable-2023, abdelnabi-adverasarial-2021}. Such methods are useful if the model is known.

\textbf{Statistical methods} (including stylistic analysis) rely on features extracted from the text. For example, in \cite{opara-distinguishing-2024} the features from 6 categories are taken into account, e.g. lexical features (word count, char count, etc.) or named entity (first person count, direct address count, etc.). The study \cite{opara-distinguishing-2025} showed that this set of features is highly correlated with our cognitive processes and may be used to distinguish between human-written and AI-generated content. In \cite{kumarage-stylometric-2023} the authors use the following groups of features to detect AI-generated tweets: 1) Phraseology – features which quantify how the author organizes words and phrases when creating a piece of text (e.g., avg. word, sent. count, etc.), 2) Punctuation – features to quantify how the author utilizes different punctuation (e.g., avg. unique punctuation count) and 3) Linguistic Diversity – features to quantify how the author uses different words in the writing (e.g., richness and readability scores). While statistical methods offer reliability and interpretability, they often struggle with broader applicability due to their reliance on pre-defined feature sets.

\textbf{Machine learning} algorithms do not involve an explicit feature extraction step, as described in the previous sections. The classifier is given the entire text as input and must learn, as part of the training process, which characteristics of the text differ between the classes. In \cite{kadiyala-robust-2025} the authors propose a system which is based on XLM-longformer with CRF layer. In \cite{prova-detecting-2024} XGB-classifier and SVM are applied for this task. LLM-DetectAIve \cite{abassy-llm-2024} uses fine-tuned RoBERTa and DeBERTa to distinguish between four categories: (i) human-written, (ii) machine-generated, (iii) machine-written, then machine-humanized, and (iv) human-written, then machine-polished.

Recently, a number of \textbf{zero-shot methods} were proposed. The main idea is to evaluate the average per-token log probability of the generated text and thresholding \cite{solaiman-release-2019}. DetectGPT \cite{mitchell-detect-2023} uses a property of the structure of an LLM’s probability function. GPT-Who ~\cite{venkatraman-gpt-2024} employs the Uniform Information Density (UID) principle, assuming that humans prefer to spread information evenly during language production.

Given the growing significance of this field, numerous \textbf{academic competitions} have been established to assess progress. SemEval-2024 Task 8: Multidomain, Multimodel and Multilingual
Machine-Generated Text Detection \cite{wang-semeval-2024} featured three subtasks: (1) Human vs. Machine Classification – the goal of this subtask is to accurately classify a text as either produced by a human or generated by a machine; (2) Multi-Way Generator Detection aims to pinpoint the exact source of a text, i.e., determine whether it originated from a human or a specific LLM; (3) Changing Point Detection – the goal is to precisely identify the exact boundary (changing point) within a text at which the authorship transitions from a human to machine happens. The RuATD Shared Task 2022 \cite{Shamardina2022FindingsOT} was developed for Russian language and consisted of two sub-tasks: (1) to determine if a given text is automatically generated or written by a human; (2) to identify the author of a given text.
This evaluation is designed specifically to detect AI-generated scientific texts in Russian. GenAI Content Detection Task 1: English and Multilingual Machine-Generated Text Detection: AI vs. Human \cite{wang-genai-2025} has two subtasks: monolingual (English) and multilingial, where the data comes from more than 8 different domains, e.g. scientific papers, social media, emails, etc. DAGPap22 shared task \cite{kashnitsky-overview-2022} concentrates on the detection of AI-generated scientific papers. The ALTA shared tasks \cite{molla-overview-2023} aims to discriminate between human-written and synthetic text generated by LLM.

\section{Dataset}
\subsection{Text generation}

The overall corpus includes the texts from six generators, i.e., one human writer and five different LLMs.The human-written abstracts were parsed and cleaned from the digital scientific journals in Russian. The domains were the following: Math, Philology, Physics, Chemistry, Pedagogy, IT, Law, Medicine, Oil and Gas, Management, Economics, Biology. It is worth noting that Economics and Biology domains were not presented in train and dev sets but only appears in the testing stage.

For each human-written text, the models were prompted to generate the abstract based on the title and keywords. The prompt was as follows:  

\selectlanguage{russian}
\texttt{Сгенерируй краткое содержание научной статьи по заголовку и ключевым словам. Напиши только текст аннотации. Не начинай текст аннотации с фразы "В данной статье".\\ 
Заголовок: \{title\}.\\
Ключевые слова: \{keywords\}
}

We prompted each model with the same prompt without changing it.The following models were selected to generate abstracts: GPT-4-Turbo \cite{openai-gpt4-2023}, Gemma2-27B \cite{riviere-gemma-2024}, Llama3.3-70B \cite{grattafiori-llama-2024}, Deepseek-V3 \cite{liu-deepseek-2024} and GigaChat-Lite \cite{gigachat}.

The post-processing stage includes removing the LLM artifacts such as specific prefixes or inappropriate output. Also, we noticed that the models tend to begin the generation with specific patterns, so we implemented some heuristics to change the beginnings while preserving the main content of the abstracts as is.

Thus, we invite participants to propose solutions to the following key challenges:
\begin{enumerate}
    \item Handling data that extend beyond the training set (generalization to new domains).
    \item Detecting texts generated by a model not included in the training data (generalization to new models).
\end{enumerate}

\subsection{Dataset overview}

The dataset size is 52,305 samples, having 35,158 samples for train, 10,978 samples for dev and 6,169 samples for test. The distribution of labels within the subsets is uniform.

The quantitative analysis of the training subset reveals several findings regarding abstract length. Human-written abstracts are significantly longer, averaging 126.4 words. In contrast, model-generated abstracts are considerably shorter, ranging from an average of 49.6 tokens for GigaChat-Lite to 85.7 tokens for GPT-4-Turbo. It is worth noting that IT and Philology are the domains with the longest human-written abstracts. Another interesting finding is that humans, Llama-3.3 and GPT-4-Turbo have the closest to the average number of words in the sentence, while Gemma2-27B and GigaChat-Lite generate sentences with a fewer number of words. Another distinguishing feature is usage of digitals – humans append in 10 times more digits in the texts than the models.

\section{Task organization}

The general purpose of the competition is to identify the precise origin of a given text, determining if it was authored by a human or generated by a particular large language model. The texts are abstracts of the scientific papers in Russian.

The shared task was run in two phases:

\textbf{Development phase}. The training and development data were available to the participants. 
The training data contained the texts and the corresponding labels reflecting the author of the text: human, GPT4-Turbo, Gemma2-27B and Llama3.3-70B. These data are assumed to be used to develop the system.
The development data contained additional generations from the unseen model which was GigaChat-Lite. The participants didn’t know the ground truth labels for the development set but they could submit the results on Codalab and get the results. We didn't limit the number of submission during this stage.
The leaderboard showed the best results for each participating team, regardless of the submission time.

\textbf{Test phase}. To assess the system's ability to generalization during the test phase both new domains (Economics and Biology) and generator (DeepSeek) were presented. This private set contained only the texts, without ground truth labels.
The duration of this stage was one week. The participants had only 5 attempts to submit their results. The number of submissions was limited to avoid overfitting on the test data.
The submission with the highest score was considered to be the final team's result.

After the competition ended, we released the gold labels for both the development and test sets. Furthermore, we kept the submission system open for the test dataset for post-shared task evaluations.

\section{Evaluation and results}

We received 159 submissions from ten teams in the development stage and five teams in the test stage. Accuracy was used to evaluate the submissions, as in the RuATD Shared Task 2022~\cite{Shamardina2022FindingsOT}. 
Table~\ref{table:dev_res} shows the scores on the development set for all submitted systems. The top two systems on the development set—GigaCheck (Mistral-7B) and YandexGPT 8B—demonstrated a clear performance advantage over traditional methods and baselines.

\begin{table}[!ht]
\centering
\begin{tabular}{l|c|c|l}
\toprule
\textbf{User}  & \textbf{Entries} & \textbf{Accuracy} & System Summary \\
\midrule
sastsy & 29 & \textbf{0.9122} & GigaCheck (Mistral-7B) \\
adugeen & 64 & 0.8696 & YandexGPT 8B \\
Nick & 13 & 0.8245 & n/a \\
vikosik3000 & 3 & 0.8164 & n/a \\
\textit{Baseline} & - & \textit{0.8081} & \textit{LogReg + TF-IDF} \\
kelijah & 2 & 0.8068 & n/a \\
fedrshm & 10 & 0.7999 & n/a \\
\textit{Baseline} & - & \textit{0.7903} & \textit{BERT} \\
chrnegor & 12 & 0.7833
 & n/a \\
dorj & 5 & 0.7564 & n/a \\
FedorinovVladislav & 2 & 0.6468 & n/a \\
eborisov & 1 & 0.2009 & n/a \\

\bottomrule
\end{tabular}

\caption{Accuracy on the development set. The best results are in bold.}
\label{table:dev_res}
\end{table}

Team \textbf{sastsy} leveraged Mistral-7B-v0.3 as its backbone LLM, enhanced with a dual-head architecture. The first head performs binary classification (Human vs. AI), while the second identifies specific AI models (e.g., GPT-4, LLaMA-3) through multiclass classification. To optimize training efficiency, the model employs LoRA (Low-Rank Adaptation), enabling lightweight fine-tuning with minimal parameter updates. Additionally, a weighted cross-entropy loss is used to ensure balanced learning across imbalanced datasets, thereby improving detection accuracy. Further implementation details are described in the paper~\cite{tolstykh2024gigacheck}.

Team \textbf{adugeen} applied a combined approach based on statistical and neural model features to improve overall detection performance. In such a hybrid architecture, the linear layers and the classifier were fine-tuned, while the rest of the model was kept frozen. Fine-tuning the YandexGPT 8B model yielded the best performance on the development set. On the test set, the best results were achieved using a combination of bag-of-words features and binoculars derived from the Gemma 2B and LLaMA 1B models. This team submitted the highest number of entries to the leaderboard.

Overall, the final results on the test set (see Table~\ref{table:test_res}) further emphasize the growing dominance of large language models in achieving high accuracy in the detection of AI-generated scientific abstracts in Russian.

\begin{table}[!ht]
\centering
\begin{tabular}{l|c|c|l}
\toprule
\textbf{User}  & \textbf{Entries} & \textbf{Accuracy} & System Summary \\
\midrule
sastsy & 3 & \textbf{0.8635} & GigaCheck (Mistral-7B) \\
adugeen & 4 & 0.8462 & BoW + b-Gemma 2B + b-Llama 1B \\
vikosik3000 & 1 & 0.8159 & n/a \\
\textit{Baseline} & - & \textit{0.8105} & \textit{LogReg + TF-IDF} \\
ESBaklanova & 1 & 0.2099 & n/a \\
fedrshm & 9 & 0.2072 & n/a \\

\bottomrule
\end{tabular}

\caption{Accuracy on the test set. The best results are in bold.}
\label{table:test_res}
\end{table}

\section{Limitations}

Due to resource limits, we only generated  abstracts. The more challenging task is to generate the whole paper text. 
Another limitation is that the generation was performed based only on the title and keywords. However, using the text or other metadata could improve the generated texts.
Also the task was limited by the classification whether the whole text of abstract is generated or not. But in practice, the most general case is to edit the generated text or to generate the more proof-read version of the human texts. The task of AI-generated spans detection is considered as one of the future research direction.

\section{Conclusion}
We presented the AINL-Eval 2025 Shared Task, focused on the detection of AI-generated scientific abstracts in Russian. The best solution for the shared task achieved 91.22\% accuracy on the development set and 86.35\% accuracy on the test set.
By introducing a comprehensive, multi-domain, and multi-model dataset, we have provided a specialized testbed for evaluating the robustness and generalizability of AI-generated text detection systems. The task design, which explicitly challenged participants to handle unseen domains and models, pushed the boundaries of current detection capabilities and highlighted the need for more sophisticated and adaptable solutions.


\begin{thebibliography}{10}

\bibitem{Ayache2022Generating}
Eliot H. Ayache and Conor M.B. Omand, \emph{Generating Scientific Articles with Machine Learning}, arXiv preprint arXiv:2203.16569, 2022.

\bibitem{chen-etal-2021-scixgen-scientific}
Hong Chen, Hiroya Takamura, and Hideki Nakayama, \emph{SciXGen: A Scientific Paper Dataset for Context-Aware Text Generation}. In Findings of the Association for Computational Linguistics: EMNLP 2021. Punta Cana, Dominican Republic. Association for Computational Linguistics. pages 1483–1492, 2021.

\bibitem{Xu2021Neural}
Sheng Xu and Xiaojun Wan, \emph{Neural Content Extraction for Poster Generation of Scientific Papers}, arXiv preprint arXiv:2112.08550, 2021.

\bibitem{Gu2024Interesting}
Xuemei Gu and Mario Krenn, \emph{Interesting Scientific Idea Generation using Knowledge Graphs and LLMs: Evaluations with 100 Research Group Leaders}, arXiv preprint arXiv:2405.17044, 2024.

\bibitem{Radensky2024Scideator}
Marissa Radensky, Simra Shahid, Raymond Fok, Pao Siangliulue, Tom Hope and Daniel S. Weld, \emph{Scideator: Human-LLM Scientific Idea Generation Grounded in Research-Paper Facet Recombination}, arXiv preprint arXiv:2409.14634, 2024.

\bibitem{Si2024Can}
Chenglei Si, Diyi Yang and Tatsunori Hashimoto, \emph{Can LLMs Generate Novel Research Ideas? A Large-Scale Human Study with 100+ NLP Researchers}, arXiv preprint arXiv:2409.04109, 2024.

\bibitem{abassy-etal-2024-llm}
Mervat Abassy, Kareem Elozeiri, Alexander Aziz, Minh Ngoc Ta, Raj Vardhan Tomar, Bimarsha Adhikari, Saad El Dine Ahmed, Yuxia Wang, Osama Mohammed Afzal, Zhuohan Xie, Jonibek Mansurov, Ekaterina Artemova, Vladislav Mikhailov, Rui Xing, Jiahui Geng, Hasan Iqbal, Zain Muhammad Mujahid, Tarek Mahmoud, Akim Tsvigun, Alham Fikri Aji, Artem Shelmanov, Nizar Habash, Iryna Gurevych, and Preslav Nakov, \emph{LLM-DetectAIve: a Tool for Fine-Grained Machine-Generated Text Detection}. In Proceedings of the 2024 Conference on Empirical Methods in Natural Language Processing: System Demonstrations, pages 336–343, Miami, Florida, USA. Association for Computational Linguistics. 2024.

\bibitem{wang-etal-2024-m4}
Yuxia Wang, Jonibek Mansurov, Petar Ivanov, Jinyan Su, Artem Shelmanov, Akim Tsvigun, Chenxi Whitehouse, Osama Mohammed Afzal, Tarek Mahmoud, Toru Sasaki, Thomas Arnold, Alham Aji, Nizar Habash, Iryna Gurevych, Preslav Nakov, \emph{M4: Multi-generator, Multi-domain, and Multi-lingual Black-Box Machine-Generated Text Detection}, Proceedings of the 18th Conference of the European Chapter of the Association for Computational Linguistics (Volume 1: Long Papers), pages 1369-1407, Association for Computational Linguistics, 2024.

\bibitem{Shamardina2022FindingsOT}
Tatiana Shamardina, Vladislav Mikhailov, Daniil Chernianskii, Alena Fenogenova, Marat Saidov, Anastasiya Valeeva, Tatiana Shavrina, Ivan Smurov, Elena Tutubalina, and Ekaterina Artemova, \emph{Findings of the The RuATD Shared Task 2022 on Artificial Text Detection in Russian}, arXiv preprint arXiv:2206.01583, 2022.

\bibitem{pmlr-v202-kirchenbauer23a}
John Kirchenbauer, Jonas Geiping, Yuxin Wen, Jonathan Katz, Ian Miers and Tom Goldstein. \emph{A Watermark for Large Language Models}. In Proceedings of Machine Learning Research, pages 17061-17084, 2023.

\bibitem{bi-2007-robust}
Ning Bi, Qiyu Sun, Daren Huang, Zhihua Yang, and Jiwu Huang. \emph{Robust Image Watermarking Based on Multiband Wavelets and Empirical Mode Decomposition}. Trans. Img. Proc. 16, 8 (August 2007), pages 1956–1966, 2007.

\bibitem{wen-tree-2023}
Yuxin Wen, John Kirchenbauer, Jonas Geiping, and Tom Goldstein. \emph{Tree-rings watermarks: invisible fingerprints for diffusion images}. In Proceedings of the 37th International Conference on Neural Information Processing Systems (NIPS '23). Curran Associates Inc., Red Hook, NY, USA, Article 2529, pages 58047–58063, 2023.

\bibitem{fernandez-stable-2023}
Pierre Fernandez, Guillaume Couairon, Hervé Jégou, Matthijs Douze and Teddy Furon. \emph{The Stable Signature: Rooting Watermarks in Latent Diffusion Models}. ICCV 2023 - International Conference on Computer Vision, Oct 2023, Paris, France, pages 22409-22420, 2023.

\bibitem{abdelnabi-adverasarial-2021}
Sahar Abdelnabi and Mario Fritz. \emph{Adversarial Watermarking Transformer: Towards Tracing Text Provenance with Data Hiding}. 42nd IEEE Symposium on Security and Privacy, pages 121-140, 2021.

\bibitem{opara-distinguishing-2024}
Chidimma Opara. \emph{StyloAI: Distinguishing AI-Generated Content with Stylometric Analysis}. In Proceedings of 25th International Conference on Artificial on Artificial Intelligence in Education(AIED 2024), 2024.

\bibitem{opara-distinguishing-2025}
Chidimma Opara. \emph{Distinguishing AI-Generated and Human-Written Text Through Psycholinguistic Analysis
}, arXiv preprint arXiv:2505.01800, 2025.

\bibitem{kumarage-stylometric-2023}
Tharindu Kumarage, Joshua Garland, Amrita Bhattacharjee, Kirill Trapeznikov, Scott Ruston and Huan Liu. \emph{Stylometric Detection of AI-Generated Text in Twitter Timelines}, arXiv preprint arXiv:2303.03697, 2023.

\bibitem{kadiyala-robust-2025}
Ram Mohan Rao Kadiyala, Siddartha Pullakhandam, Kanwal Mehreen, Drishti Sharma, Siddhant Gupta, Jebish Purbey, Ashay Srivastava, Subhasya TippaReddy, Arvind Reddy Bobbili, Suraj Telugara Chandrashekhar, Modabbir Adeeb, Srinadh Vura and Hamza Farooq. \emph{Robust and Fine-Grained Detection of AI Generated Texts}, arXiv preprint arXiv:2504.11952, 2025.

\bibitem{prova-detecting-2024}
Nuzhat Prova. \emph{Detecting AI Generated Text Based on NLP and Machine Learning Approaches}. arXiv preprint arXiv:2404.10032, 2024.

\bibitem{abassy-llm-2024}
Mervat Abassy, Kareem Elozeiri, Alexander Aziz, Minh Ngoc Ta, Raj Vardhan Tomar, Bimarsha Adhikari, Saad El Dine Ahmed, Yuxia Wang, Osama Mohammed Afzal, Zhuohan Xie, Jonibek Mansurov, Ekaterina Artemova, Vladislav Mikhailov, Rui Xing, Jiahui Geng, Hasan Iqbal, Zain Muhammad Mujahid, Tarek Mahmoud, Akim Tsvigun, Alham Fikri Aji, Artem Shelmanov, Nizar Habash, Iryna Gurevych and Preslav Nakov. \emph{LLM-DetectAIve: a Tool for Fine-Grained Machine-Generated Text Detection}. In Proceedings of the 2024 Conference on Empirical Methods in Natural Language Processing: System Demonstrations, pages 336–343, 2024.

\bibitem{solaiman-release-2019}
Irene Solaiman, Miles Brundage, Jack Clark, Amanda Askell, Ariel Herbert-Voss, Jeff Wu, Alec Radford, Gretchen Krueger, Jong Wook Kim, Sarah Kreps, Miles McCain, Alex Newhouse, Jason Blazakis, Kris McGuffie and Jasmine Wang. \emph{Release Strategies and the Social Impacts of Language Models}. arXiv preprint arXiv:1908.09203, 2019.

\bibitem{mitchell-detect-2023}
Eric Mitchell, Yoonho Lee, Alexander Khazatsky, Christopher D. Manning, and Chelsea Finn. \emph{DetectGPT: zero-shot machine-generated text detection using probability curvature}. In Proceedings of the 40th International Conference on Machine Learning (ICML'23), Vol. 202, pages 24950–24962, 2023.

\bibitem{venkatraman-gpt-2024}
Saranya Venkatraman, Adaku Uchendu and Dongwon Lee. \emph{GPT-who: An Information Density-based Machine-Generated Text Detector}. In Findings of the Association for Computational Linguistics: NAACL 2024, pages 103–115, 2024.

\bibitem{wang-semeval-2024}
Yuxia Wang, Jonibek Mansurov, Petar Ivanov, Jinyan Su, Artem Shelmanov, Akim Tsvigun, Osama Mohammed Afzal, Tarek Mahmoud, Giovanni Puccetti and Thomas Arnold. \emph{SemEval-2024 Task 8: Multidomain, Multimodel and Multilingual Machine-Generated Text Detection}. In Proceedings of the 18th International Workshop on Semantic Evaluation (SemEval-2024), pages 2057–2079, 2024.

\bibitem{wang-genai-2025}
Yuxia Wang, Artem Shelmanov, Jonibek Mansurov, Akim Tsvigun, Vladislav Mikhailov, Rui Xing, Zhuohan Xie, Jiahui Geng, Giovanni Puccetti, Ekaterina Artemova, Jinyan Su, Minh Ngoc Ta, Mervat Abassy, Kareem Ashraf Elozeiri, Saad El Dine Ahmed El Etter, Maiya Goloburda, Tarek Mahmoud, Raj Vardhan Tomar, Nurkhan Laiyk, Osama Mohammed Afzal, Ryuto Koike, Masahiro Kaneko, Alham Fikri Aji, Nizar Habash, Iryna Gurevych and Preslav Nakov. \emph{GenAI Content Detection Task 1: English and Multilingual Machine-Generated Text Detection: AI vs. Human}. In Proceedings of the 1stWorkshop on GenAI Content Detection (GenAIDetect), pages 244–261, 2025.

\bibitem{kashnitsky-overview-2022}
Yury Kashnitsky, Drahomira Herrmannova, Anita de Waard, George Tsatsaronis, Catriona Catriona Fennell and Cyril Labbe. \emph{Overview of the DAGPap22 Shared Task on Detecting Automatically Generated Scientific Papers}. In Proceedings of the Third Workshop on Scholarly Document Processing, pages 210–213, 2022.

\bibitem{molla-overview-2023}
Diego Molla, Haolan Zhan, Xuanli He and Qiongkai Xu. \emph{Overview of the 2023 ALTA Shared Task: Discriminate between Human-Written and Machine-Generated Text}. In Proceedings of the 21st Annual Workshop of the Australasian Language Technology Association, pages 148–152, 2023.

\bibitem{openai-gpt4-2023}
OpenAI, Josh Achiam, Steven Adler, Sandhini Agarwal, Lama Ahmad, Ilge Akkaya, Florencia Leoni Aleman et al. \emph{GPT-4 Technical Report}. arXiv preprint arXiv:2303.08774, 2023.

\bibitem{riviere-gemma-2024}
Morgane Riviere, Shreya Pathak, Pier Giuseppe Sessa, Cassidy Hardin, Surya Bhupatiraju, Léonard Hussenot, Thomas Mesnard, Bobak Shahriari, Alexandre Ramé, Johan Ferret et al. \emph{Gemma 2: Improving Open Language Models at a Practical Size}. arXiv preprint 	arXiv:2408.00118, 2024.

\bibitem{grattafiori-llama-2024}
Aaron Grattafiori, Abhimanyu Dubey, Abhinav Jauhri, Abhinav Pandey, Abhishek Kadian, Ahmad Al-Dahle, Aiesha Letman, Akhil Mathur, Alan Schelten, Alex Vaughan et al. \emph{The Llama 3 Herd of Models}. arXiv preprint arXiv:2407.21783, 2024.

\bibitem{liu-deepseek-2024}
Aixin Liu, Bei Feng, Bing Xue, Bingxuan Wang, Bochao Wu, Chengda Lu, Chenggang Zhao, Chengqi Deng, Chenyu Zhang, Chong Ruan, Damai Dai, Daya Guo, Dejian Yang, Deli Chen et al. \emph{DeepSeek-V3 Technical Report}. arXiv preprint arXiv:2412.19437, 2024.

\bibitem{gigachat}
Valentin Mamedov, Evgenii Kosarev, Gregory Leleytner, Ilya Shchuckin, Valeriy Berezovskiy, Daniil Smirnov, Dmitry Kozlov et al. \emph{GigaChat Family: Efficient Russian Language Modeling Through Mixture of Experts Architecture}. In Proceedings of the 63rd Annual Meeting of the Association for Computational Linguistics. Volume 3: System Demonstrations, pages 93-106, 2025.

\bibitem{tolstykh2024gigacheck}
Irina Tolstykh, Aleksandra Tsybina, Sergey Yakubson, Aleksandr Gordeev, Vladimir Dokholyan, and Maksim Kuprashevich, \emph{GigaCheck: Detecting LLM-generated Content}, arXiv preprint arXiv:2410.23728, 2024.

\end{thebibliography}
\end{document}